\documentclass[journal]{IEEEtran}

\IEEEoverridecommandlockouts

\usepackage{amsthm}
\usepackage{subfiles}
\usepackage{wrapfig}
\usepackage{lipsum}
\usepackage{footnote}
\usepackage{amssymb} 
\usepackage{yfonts}
\usepackage{float}
\usepackage{comment}

\usepackage[english]{babel}
\newtheorem{definition}{Definition}
\newtheorem{lemma}{Lemma}
\newtheorem{theorem}{Theorem}
\newcounter{subdefinition}[definition]
\renewcommand{\thesubdefinition}{\thedefinition.\arabic{subdefinition}}

\usepackage{booktabs}
\usepackage{algorithm}
\usepackage{multirow}
\usepackage{algpseudocode}
\usepackage{graphicx}
\usepackage{stfloats}
\usepackage{tabularx}
\usepackage{arydshln} 
\usepackage[export]{adjustbox}

\usepackage{amsmath}
\usepackage[inkscapelatex=false]{svg}
\usepackage{subcaption}
\usepackage{avant}

\usepackage[T1]{fontenc}

\usepackage[colorlinks=true]{hyperref}
\usepackage[multiple]{footmisc}

\newtheoremstyle{named}{}{}{\itshape}{}{\bfseries}{.}{.5em}{\thmnote{#3}#1}
\theoremstyle{named}

\title{\textbf{KT-BT: A Framework for Knowledge Transfer Through Behavior Trees in Multi-Robot Systems}}

\author{Sanjay Sarma O V$^{1}$ \and Ramviyas Parasuraman$^{2,*}$ \and Ramana Pidaparti$^{3}$
\thanks{$^{1}$School of Electrical and Computer Engineering, University of Georgia, Athens, GA 30602, USA. 
        email: {\tt\small sanjaysarmaov@uga.edu}.}%
\thanks{$^{2}$School of Computing, University of Georgia, Athens, GA 30602, USA.  
        email: {\tt\small ramviyas@uga.edu}  $^*$Corresponding author. }%
\thanks{$^{3}$School of Environmental, Civil, Agricultural and Mechanical Engineering, University of Georgia, Athens, GA 30602, USA. 
        {\tt\small rmparti@uga.edu}}%
}

\begin{document}
\maketitle

\begin{abstract}
Multi-Robot and Multi-Agent Systems demonstrate collective (swarm) intelligence through systematic and distributed integration of local behaviors in a group. Agents sharing knowledge about the mission and environment can enhance performance at individual and mission levels. However, this is difficult to achieve, partly due to the lack of a generic framework for transferring part of the known knowledge (behaviors) between agents. This paper presents a new knowledge representation framework and a transfer strategy called KT-BT: Knowledge Transfer through Behavior Trees. The KT-BT framework follows a query-response-update mechanism through an online Behavior Tree framework, where agents broadcast queries for unknown conditions and respond with appropriate knowledge using a condition-action-control sub-flow. We embed a novel grammar structure called \textit{stringBT} that encodes knowledge, enabling behavior sharing. We theoretically investigate the properties of the KT-BT framework in achieving homogeneity of high knowledge across the entire group compared to a heterogeneous system without the capability of sharing their knowledge. We extensively verify our framework in a simulated multi-robot search and rescue problem. The results show successful knowledge transfers and improved group performance in various scenarios. We further study the effects of opportunities and communication range on group performance, knowledge spread, and functional heterogeneity in a group of agents, presenting interesting insights.
\end{abstract}


\IEEEpeerreviewmaketitle

\section{INTRODUCTION}

Humans and animals developed social communication as an evolutionary trait over thousands of years to help each other locate potential food opportunities, predators, migratory information, etc. \cite{Kaplan2014AnimalCommunication}. These communications are ubiquitous and are crucial for decision-making under unknown circumstances and determine both the individual and group's survival and benefits \cite{RendallDSignallers}. This may involve different auditory, visual, olfactory, or tactile communication modalities and their combinations depending on the type of information or knowledge transmitted. 

Knowledge and information sharing strategies similar to those in natural systems have also been studied and applied in information science, multi-agent systems (MAS), machine learning, and IoT focused on developing collective intelligence \cite{Whiten2022TheMachines}. 

Many important pieces of information are combined to generate knowledge from which inferences, action sequences, and predictions are made. While information is directly transferable, knowledge sharing may sometimes involve learning and require unique ways to transfer.

\begin{figure}[t]
    \centering
    \includegraphics[width=\linewidth]{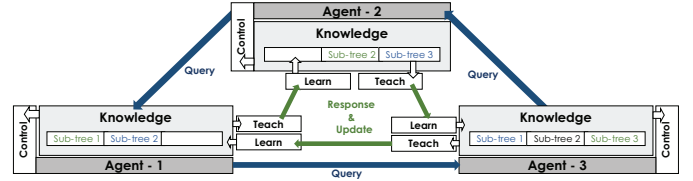}
    \caption{An overview of Multi-Agent Knowledge Transfer through Behavior Trees (KT-BT) framework, where learners and teachers use a query-response-update style with BTs.}
    \vspace{-4mm}
    \label{fig:MATBT_Overview}
   
\end{figure}

In robotics, like in many animals, the knowledge is procedural that defines the robot's ability to perform a given task by synthesizing information from sensory data \cite{riedl2019human}. It can be the knowledge of performing a routine assembly operation performed in automobile manufacturing or cleaning a carpeted floor \cite{Hautala2021CanFactory}. This knowledge can either be acquired through various learning strategies like observation and imitation \cite{fitzgerald2014visual}, using machine learning \cite{YongyuanLiang2020ParallelLearning}, and transfer learning \cite{fitzgerald2016situated}, knowledge sharing through query-response mechanisms \cite{Florez-Puga2009Query-enabledTrees}, or pre-programmed. In many cases, the knowledge is limited or unique to a single robot or may involve multi-robot learning, learning from humans through demonstration or clarification, etc. \cite{racca2019teacher,ravichandar2020recent}. In the case of a multi-robot system, a common knowledge-sharing framework is currently lacking but is highly critical to facilitate robust knowledge transfer and ensure mission performance at the group level. 

With the increasing pervasiveness of robots in industries \cite{mills1996dynamic, touzani2021multi}, agriculture \cite{r2018research, mao2021research}, transportation \cite{badue2021self}, defense \cite{rouvcek2019darpa, gans2021cooperative}, and security, the centralization of knowledge is sometimes complex and challenging. Additionally, direct information exchange between the agents has advantages and enhances the system-level performance and robustness while also reducing the design complexities \cite{sharma2016comparative}. This is very significant in mission control involving a variety of robots, where seamless transfer of knowledge requires a common acceptable framework across both homogeneous and heterogeneous groups in a multi-robot system (MRS) \cite{parasuraman2020impact,van2007ontology, kono2014transfer}.

To remedy these gaps, we draw inspiration from social communication in natural systems and propose a new framework called Knowledge Transfer Behavior Trees (KT-BT). Fig.~\ref{fig:MATBT_Overview} depicts an overview of the KT-BT framework. KT-BT is built over Behavior Trees (BTs), which are historically applied to automate non-player characters in games \cite{sekhavat2017behavior} and gathered recent applications to robotics and AI \cite{colledanchise2018behavior,iovino2022survey}. 
The framework encapsulates knowledge in a hierarchical structure containing various sub-trees, each representing a particular knowledge (task-level condition-action tuples) that can be explicitly shared across the multi-robot system. 

The main contributions of this paper are summarized below.
\begin{enumerate}
    \item We define a novel query-based knowledge sharing framework called KT-BT, where robots\footnote{We use the term "robot" and "agent" interchangeably to represent an autonomous agent with some intelligence.} explicitly communicate and collaborate to share parts of their exclusive knowledge base\footnote{The knowledge is a set of functional skills a robot possesses to execute an action plan based on the current state. The knowledge base of each robot could be pre-programmed or learned through some techniques like Reinforcement Learning, but we assume they are pre-programmed (encoded) in their high-level state-action planning framework. Further, to simplify the concepts, we assume that each robot has some part of the knowledge that is commonly present on all robots and some part of the knowledge that is unique to a robot.}. 

    \item We utilize the modularity and hierarchy features of BTs in designing a query-response mechanism among multiple agents and embed a new mechanism to incorporate the updated knowledge by recompiling the BTs at runtime, enabling the framework to work online for robot control and collaboration.

    \item We introduce a unique BT representation using novel grammar constructs called \textit{stringBT} that enables the protocols for query, quick retrieval of sub-trees, and response for achieving knowledge sharing.

    \item We theoretically analyze the properties of KT-BT in guaranteeing knowledge transfer between robots, increasing knowledge spread across the group, and enhancing opportunities to improve mission performance in a generic multi-agent framework. 
    \item We present an application of the KT-BT framework on a Search and Rescue (SAR) problem simulation involving multiple robots\footnote{We released an executable version of the simulator at \url{https://github.com/herolab-uga/KTBT-Release} to provide the readers a sandbox platform to configure the SAR simulation settings and verify the advantages of KT-BT.}. Here, we validate the advantages and demonstrate the utility of the KT-BT framework in terms of homogenizing the knowledge spread and improving the overall mission performance and efficiency.
\end{enumerate}

Finally, we organize the remainder of the paper as follows. Sec.~\ref{sec:relatedwork} briefly reflects on the knowledge-sharing strategies proposed in the literature. In Sec.~\ref{sec:KTBT}, we discuss some background on behavior trees and formally introduce our KT-BT framework through definitions, architectures, and algorithms. Sec~\ref{sec:formulation} presents the theorems that establish the knowledge spread in a multi-agent group. We validate the KT-BT framework on an application considering a multi-robot search and rescue (SAR) or multi-target foraging problem in Sec.~\ref{sec:SAR}. The results and analyses on knowledge propagation from the simulations are presented in Sec.~\ref{sec:experiment}. Finally, Sec.~\ref{sec:conclusion} concludes the paper.

\section{Related Work}
\label{sec:relatedwork}
 
We present an overview of various ontologies for knowledge representation and sharing in autonomous agents and discuss how our KT-BT framework differs from state-of-the-art. 

The transfer of information via any modality like an Agent Communication Language \cite{Soon2019ALanguage} generally involves identifying queries, responding appropriately, and merging the response with the existing knowledge. Hence, knowledge should be represented so that it is easily accessible, retrievable, and shared with the group \cite{Tamma2002AnSystems} for individual and collective decision-making 
\cite{yang2022game}.  Moreover, the transfer of information is easier to achieve than the transfer of knowledge itself.

In this regard, ontologies have gained a reputation for their flexibility and robustness in knowledge representation as they are designed to be unambiguous that can be reused, fragmentized, or directly shared with other agents \cite{Dorri2018Multi-AgentSurvey}. In ontologies, vocabulary for concepts \cite{Gruber1993ASpecifications} are defined along with their relationships and constraints in the form of axioms. Their logic is described using syntax and semantics, and operations like merging, mapping, alignment, unification, refinement, and inheritance are performed on the relationship maps to dynamically update ontologies with new knowledge \cite{staab2010handbook}.

Research in MRS has seen many robust applications of ontologies for inter-robot knowledge transfer. The standards like CORA (Core Ontologies for Robotics and Automation) \cite{schlenoff2012ieee} accelerated the development of knowledge sharing in both homogeneous and heterogeneous multi-robot teams and primarily focused on human-robot interactions, positioning systems, or industrial settings \cite{Olszewska2017OntologyRobotics, Fiorini2017AActivities}. 

In \cite{Skarzynski2018SO-MRS:Ontology}, the authors presented a service-oriented architecture called SO-MRS for heterogeneous multi-robot communication, which exemplifies the standard strategies of representing service requests and environment in an ontology language. Saigol et al. \cite{Saigol2013FacilitatingFramework} developed a knowledge-sharing framework between UAVs using ontologies, in which they encode ontologies onto acoustic packets that are transferred to other UAVs. 
Other examples include a cloud-based knowledge sharing mechanism in combination with Deep Reinforcement learning for optimizing the service schedules between industrial robots \cite{Du2019CollaborativeSharing}. In their work, knowledge sharing was formulated between the cloud and robots (R2C) and between robots (R2R), and the Web Ontology Language (OWL) was used for knowledge encoding. 
A similar interesting work by Chen et al. \cite{Iovino2020AAI} proposes sharing and distribution of knowledge of a robot that needs disengagement from the process due to deterioration or maintenance. This helps to maintain the new attending robot's skill capacity level as that of a retiring robot while also ensuring a good production performance in a cell.

In some hybrid techniques involving multi-agent reinforcement learning, agents combine the policies or knowledge with ontological representation for sharing with other agents. 
For example, Qu et al. \cite{Qu2016OptimizedApproach} proposed a framework with multi-agent reinforcement learning for optimal scheduling of a multi-skill workforce and multiple machines for a multi-stage manufacturing process. 
Similarly, Oprea et al. \cite{oprea2018agent} use a combination of ontologies and Q - learning for agent adaptation. Taylor et al. \cite{Taylor2019ParallelTransfer} proposed a parallel transfer learning technique where the selected knowledge is shared with agents in parallel. 

A robot receiving new knowledge shared by other robots can strategically decide on the need for merging by comparing the rewards from its experience. For instance, a confidence-based approach can be used to accommodate this knowledge \cite{chernova2010confidence}. Alternatively, a value function can be used by agents to share policies mutually, and each individual agent uses a common model to combine its expert policy with the multi-agent network policy in deducing a joint policy \cite{Liu2019ValueReturns}. 
A similar query-answer-based model-sharing was proposed by Jiang et al.\cite{jiang2020model} and Zhou et al. \cite{zhou2016multiagent}, in which an equilibrium-based sparse interaction framework that shares local Q-values with other agents called NegoSI was developed. 

In all these works involving applications of ontologies in MAS and MRS research, we can observe that the knowledge does not directly represent the control actions at a lower level or instead supports only high-level decisions and inter-robot/agent communications. We identify a lack of a unifying model framework that works at all levels combining decision making, control, knowledge sharing, and communication. Additionally, applying ontologies requires clear concept definitions, establishing relationships between concepts, and defining constraints, which requires a good amount of domain expertise and knowledge of using various tools. 

On the other hand, there is a growing interest in applying Behavior Trees (BTs) to robotics, multi-agent, and multi-robot control for their scalability, modularity, reactive, and safety guarantee properties \cite{Iovino2020AAI,Marzinotto2014TowardsControl}. A summary of various nodes and components of BT design along with an application on a humanoid robot is presented \cite{colledanchise2018behavior,giunchiglia2019conditional}. BTs have shown excellent advantages in robotics \cite{Rovida2017ExtendedTasks} and MRS \cite{colledanchise2016advantages,yang2020needs,colledanchise2018learning}. For instance, in \cite{Agis2020AnGames}, a BT-based mechanism with explicit communication requests was proposed for multi-agent event-driven coordination in non-player characters of video games.

Contrary to the existing frameworks for knowledge representation and sharing, we use BTs to represent knowledge and propose a grammar protocol for sharing knowledge. BTs are uniquely suited to our knowledge-sharing framework because they are capable of combining control, planning, and learning into a single unifying framework \cite{styrud2022combining,iovino2022survey,colledanchise2018learning}. 
Also, in comparison to ontology-based methods that are mostly for knowledge representation only, our framework using BTs provides the flexibility of knowledge representation, high-level decision-making, hierarchical state-action planning, and low-level control execution.

\begin{figure}[t]
    \centering
    \includegraphics[width=0.98\linewidth]{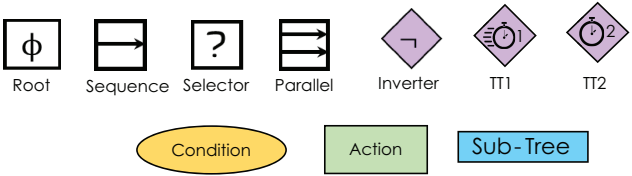}
    \caption{Various BT nodes used in the current study.}
    \label{fig:BTNodes}
    \vspace{-4mm}
\end{figure}
However, the current BT-based methods lack tools to expand knowledge-sharing between multiple agents and to perform BT operations similar to ontologies. 
In addition, there is no consensus in the literature on a standard design template for task-agnostic BT design. Our work is in the direction of representing a knowledge-based BT template that is generalizable across various tasks and applications.

Therefore, we propose the new KT-BT framework that uses a query-response mechanism for explicitly sharing knowledge using communication in MRS. To the best of our knowledge, our KT-BT framework is the first work in the literature that incorporates the new knowledge (or intelligence) in a real-time manner (through live updates of their BTs) while the robots are performing their current control actions using their current BTs. Furthermore, present some first-of-a-kind investigations on the properties of knowledge sharing and its spread in an MRS group under various scenarios in a search and rescue simulation case study. 
We believe these advances will help advance the research in robotics and MAS/MRS by enabling explicit knowledge sharing.

\begin{table}[b]
\caption{A summary of the functions of control, decorators, and execution nodes used in the current study.}
\label{table:NodesSummary}
\begin{tabular}{ll}
\toprule
\textbf{Type of Node}	& \textbf{Function} \\
\midrule
Sequence & \multicolumn{1}{m{6cm}}{Runs children nodes from left to right till a child node returns false. Returns true when all children nodes return true.}\\ 
Selector & \multicolumn{1}{m{6cm}}{Runs child nodes from left to right till a child node returns true. Returns true when at least one child node returns true.}\\
Parallel &\multicolumn{1}{m{6cm}}{Runs all children in parallel.} \\
\midrule
Inverter &\multicolumn{1}{m{6cm}}{Inverts the return value of the child.} \\
TT1  &\multicolumn{1}{m{6cm}}{Waits for a fixed number of ticks before executing the child. Returns running during the wait, and returns child return value after the wait.}\\
TT2  &\multicolumn{1}{m{6cm}}{Runs child for a fixed number of ticks. Returns child return value when running, otherwise, returns false when execution is complete.} \\
\midrule
Condition  &\multicolumn{1}{m{6cm}}{Returns true when the condition is true.} \\
Action  &\multicolumn{1}{m{6cm}}{Executes action or action sequence. Returns running during execution, true after completion.} \\
Sub-tree  &\multicolumn{1}{m{6cm}}{A smaller tree that can be merged with a larger tree.} \\
\bottomrule
\end{tabular}
\end{table}

\section{Proposed KT-BT for Knowledge Transfer Using Behavior Trees in Multi-Agent Systems}
\label{sec:KTBT}

In this section, we first present a background on BT and discuss the query-response mechanism. Then, we formulate the knowledge representation using BTs and introduce a new protocol to enable explicit query, retrieval, and sharing of part of the knowledge between robots in an MRS.

\subsection{Background on Behavior Trees}
BTs were first introduced for the control design of non-player characters (NPCs) in video games, in which the conditions and actions are mapped using control and execution nodes. They provide excellent graphical design flexibility to the user to modify the control actions and define hierarchies in task planning for agents. Over time, they found their way into robotics and other AI applications \cite{colledanchise2018behavior}.

BTs are directed trees that start with a root node and may have multiple control and execution nodes. Root nodes have no parents, execution nodes have no children, and control nodes have one parent and may have multiple children. In general, to represent BTs graphically, child nodes are represented under parent nodes, and all the execution nodes are shown as leaf nodes. Each execution of the BT happens at a certain frequency called ticks. In each tick, starting from the root node, the nodes are executed as per the control flow and from left to right. This paper follows the convention of top-down tree flow representation and left-to-right priority in execution. And thus, the high-priority nodes can be placed with the leftmost nodes that are executed at the beginning of each tick. The node representations followed in the current work are presented in Fig.~\ref{fig:BTNodes} and their summary in Table \ref{table:NodesSummary}. 

\subsubsection{Control Nodes}
A control node may have multiple children that are executed according to logic. Commonly used control nodes are selectors, sequencers, parallel, and decorators. A selector ticks children from left to right until a \textit{success} is returned by a child, and a sequencer runs all the children from left to right till a child returns a \textit{failure}. A parallel node executes all its child subtrees in parallel and generally returns a \textit{running} status \cite{colledanchise2021handling}. Finally, a decorator node is designed to modify the child's response through a policy defined by the user. 
An inverter can only have one child node and flips the return status if it is different from running. e.g., a \textit{success} is flipped as a \textit{failure}, and vice versa. 

For our work, we propose two new decorators: a $TT1$ wait timer and $TT2$ execution timer. 
A $TT1$ timer can have only one child, and it waits for a fixed number of ticks before executing the child and returns running during the wait. After the delay, it returns the child return value. Similarly, a $TT2$ executes a child for a fixed number of ticks and returns the child status during the run time and failure thereafter. 

\subsubsection{Execution nodes}
Action and condition nodes fall under the execution category, which are the leaf nodes in a BT. An action node runs an action and returns a \textit{success} if it’s completed or a \textit{failure} or \textit{running} otherwise. On the other hand, a condition node verifies if a particular condition is satisfied and returns a \textit{success} or returns a \textit{failure} otherwise. Generally, all the condition variables frequently verified through a behavior tree are maintained in a common location called a blackboard with (key, value) pairs. Similarly, in the current study, for the SAR simulations, we maintain a state manager that keeps track of all the condition variables that a Behavior Tree can access.

\subsection{Overview of query-response mechanism in KT-BTs}
In our framework, each agent has a behavior tree that defines its control, teaching (response), and learning (query and update) sequences that run in parallel. In general, each agent can exist either in a mission (executing an action using its current knowledge), teaching (responding to a query from other robots), or learning (incorporating new knowledge from other robots) modes, depending on its state and the conditions it encounters. Further, every agent's control tree consists of critical, knowledge base, and fallback sub-trees. 

While the critical and fallback sub-trees represent the agent's safety and fallback routines \cite{colledanchise2016behavior}, respectively, the knowledge sub-trees representing the agent's current knowledge base are a primary focus of our work. The agent executes the knowledge sub-trees when a specific set of conditions are met in its environment. Further, an agent also maintains a list of a known sequence of states and conditions that correspond to a new knowledge sub-tree. At any point during a mission,  the agent verifies if the encountered state and condition sequences match with the sequences corresponding to its knowledge. When an unknown sequence is encountered, the agent broadcasts a query to its neighbors, thus initiating the query-response mechanism. 

\begin{figure*}[t]
    \centering
    \includegraphics[width=\linewidth]{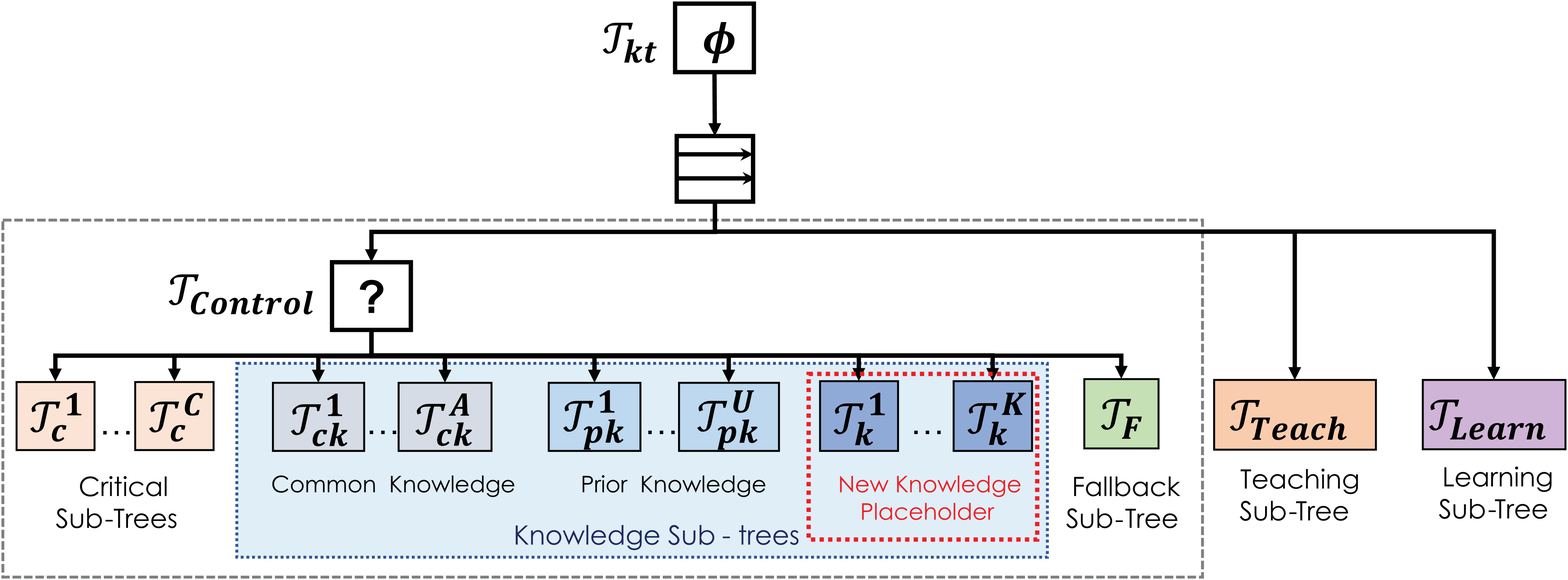}
    \caption{Overall structure of Behavior Tree designed for the current knowledge transfer study.}
    \label{fig:Def1and2}
\end{figure*}

The agent sends out the unknown sequence as a query and awaits a response. Next, a receiving agent within the querying agent's communication range verifies the query sequence against its known knowledge base and responds with the corresponding sub-tree encoded as a \textit{stringBT} (described in Sec.~\ref{sec:stringbt}). Finally, the querying agent decodes the received response and merges it with its control tree, thus continuing with the appropriate execution process. We present an overview of the knowledge transfer in our current KT-BT framework in Figs.~\ref{fig:MATBT_Overview} and \ref{fig:Def1and2}, where sub-trees are learned through query-response mechanisms between three functionally heterogeneous agents. 

This mechanism is advantageous when the agents demonstrate functional heterogeneity due to varying amounts of knowledge. For example, a team may contain only one agent with knowledge of all the tasks. With the knowledge propagating across the groups, all the agents in the MAS can develop uniform capabilities in accomplishing the low-level tasks for mission-level success by learning from this one agent who knows all tasks. Similarly, consider a scenario where each agent in an MRS group contains unique knowledge that is complementary to other agents. Exploiting a KT-BT framework, this MRS group can propagate their knowledge, and each agent will harmonize their knowledge base by combining all of their knowledge. In another example, assume a robot has the ability to learn through interaction and encode this knowledge as a BT once learned. Other agents can acquire this knowledge without having to learn on their own.

\begin{figure}[t]
    \centering
    \includegraphics[width=0.85\linewidth]{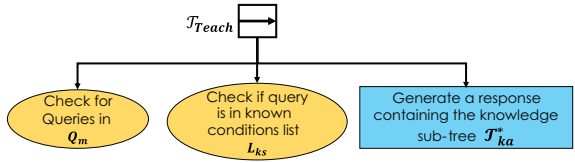}
    \caption{Teaching sub-tree consisting of a sequence node that verifies if the queried sequence $s_q$ is known and responds with the knowledge tree $\mathcal{T}_{ka}^*$ corresponding to the query $s_q$.}
    \label{fig:TeachTree1}
\end{figure}

\subsection{Knowledge formulation in KT-BTs}

A BT is defined as a three tuple, $\mathcal{T}_{lbl}^i=\{f^i,r^i,\Delta t\}_{lbl}$,  where $i \in \mathbb{N}$ is the tree index, and $lbl$ is a label that defines its class. $f^i$ is the function that maps the system's current state $s^i\in S$ to the output actions $a^i$. $\Delta t$ is a time step, and the return status is defined as $r^i:\mathbb{R}^n \xrightarrow{}\{\mathcal{R,S,F}\}$, which can either be a Running, Successful, or Failure status.
Here, we go by any assumptions and definitions of sequence and fallback (selector) as presented by Colledanchise et al., \cite{colledanchise2018behavior} in their state-space formulations for BTs.

For our current study, we designed a unique tree structure that facilitated the learning and teaching processes. We label this general tree structure as $\mathcal{T}_{kt}$ and is defined as follows.
\theoremstyle{definition}
\begin{definition}
\label{def:kt}
A transfer learning tree $\mathcal{T}_{kt}$ has three sub-trees associated with control, learning, and teaching. All these three sub-trees are run in parallel.
\begin{equation}
    \mathcal{T}_{kt}=Parallel(\mathcal{T}_{Control},\mathcal{T}_{Teach},\mathcal{T}_{Learn})
    \label{eq:kt}
\end{equation}
\end{definition}

\subsubsection{Control}
A $\mathcal{T}_{Control}$ sub-tree is divided into critical, knowledge, and fallback sub-trees, each corresponding to their intended purposes as shown in Fig.~\ref{fig:Def1and2}. For example, in the case of a mobile robot, a critical collision avoidance sub-tree with high priority is placed towards the left extreme, followed by lesser priority critical sub-trees for battery recharge or wait commands. 
Following the critical sub-trees, towards the right, are knowledge sub-trees. 

A knowledge sub-tree can be classified either into common ($\mathcal{T}_{CK}$), prior ($\mathcal{T}_{PK}$), or new knowledge ($\mathcal{T}_{K}$) sub-trees. The positions of these sub-trees may be varied depending on their order of priority. For the current framework, we maintain the priority order as common, prior, and new knowledge. A common knowledge sub-tree $\mathcal{T}_{ck}^i$in $\mathcal{T}_{CK}$ is the knowledge that is common across all the agents in an MRS group. 

In addition to common knowledge, an agent in a group may have prior knowledge $\mathcal{T}_{PK}$ that is inherent to the agent or may be acquired during a mission in the form of a new knowledge $\mathcal{T}_{K}$. We create a placeholder in each agent's $\mathcal{T}_{control}$ where this new knowledge can be placed.

Finally, the tree $\mathcal{T}_{F}$ is a set of fallback sub-trees that follow the knowledge sub-trees segment. These trees are activated when none of the conditions towards the left under the selector in $\mathcal{T}_{Control}$ are met. Some examples of fallback routines include random walk, exploration, idle/ sleep, etc. We present a formal definition of $\mathcal{T}_{Control}$ as follows.

\begin{definition}
\label{def:control}
A control sub-tree $\mathcal{T}_{Control}$ has selector with sub-trees in the order (priority) of critical sub-trees $\mathcal{T}_{C}$, action sub-trees $\mathcal{T}_{A}$, knowledge trees $\mathcal{T}_{K}$ and a fallback sub-tree $\mathcal{T}_F$.
\begin{equation}
    \mathcal{T}_{Control}=Selector(\mathcal{T}_{C},\{\mathcal{T}_{CK},\mathcal{T}_{PK},\mathcal{T}_{K}\},T_F)
    \label{eq:control}
\end{equation}
Here, $\mathcal{T}_{C}$, $\mathcal{T}_{CK}$, $\mathcal{T}_{PK}$ and $\mathcal{T}_{K}$ are ordered sets of critical, common knowledge, prior knowledge, and new knowledge sub-trees. A combined $\mathcal{T}_{Control}$ tree built from Definitions \ref{def:kt} and \ref{def:control} is shown in Fig.~\ref{fig:Def1and2}.
\end{definition}

\begin{figure}[t]
    \centering
    \includegraphics[width=0.9\linewidth]{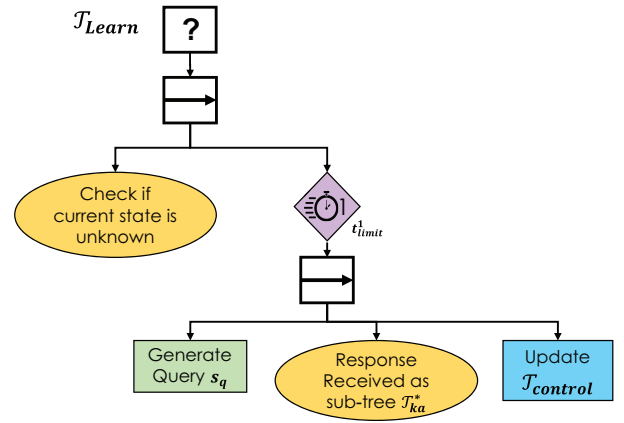}
    \caption{Learning sub-tree transmits an unknown state-sequence $s_q$ to other robots via broadcast and updates $\mathcal{T}_{Control}$ upon receiving a response $\mathcal{T}_{ka}^*$ from at least one robot.}
    \label{fig:LearnTreeRLEL}
\end{figure}

The set of knowledge sub-trees in $\mathcal{T}_{CK}, \mathcal{T}_{PK}$ and $\mathcal{T}_{K}$ are ordered. For example, a tree $\mathcal{T}_{k}^j \in \mathcal{T}_{K}$ is defined as
\begin{align}
    \mathcal{T}^{j}_{k} &=Sequence(s_{ka}^j,\mathcal{T}_{ka}^j) ,
    \label{eq:knowledge}
\end{align}
where, for a given knowledge subtree $\mathcal{T}_k^j$, $s_{ka}^j$ denotes the state sequence, which is a sequence of conditions 1 through M denoted in their sub-script $s^j = \{s_1,\dots,s_M\}^j$. $\mathcal{T}_{ka}^j$ is an action sub-tree that is run when all the conditions corresponding to the sequence in the state sequence ${s^j}$ are satisfied (for a $j^{th}$ knowledge sub-tree). 
We currently assume the conditions-actions sequence is split into two sub-trees for simplifying the analysis but they can be complexly intertwined in real applications.

\subsubsection{Teach}
\label{sec:teach}
Assume each agent maintains a list of known states $L_{ks}$ and known actions $L_{ka}$.
A teaching tree $\mathcal{T}_{Teach}$ checks for any state-sequence query $s_q$ received in a message buffer $(Q_m)$ and responds with an appropriate state-action tree $\mathcal{T}_{ka}^*$ if the state-sequence query $s_q$ is known in its knowledge base $\mathcal{T}_k$. A state-sequence query $s_q$ is considered as known if it is present in the known states list $L_{ks}$. i.e., $s_q = L_{ks}^i$, for some $i$ that maps the condition sequence to a state-action tree $\mathcal{T}_{ka}^*=\mathcal{T}_{ka}^i \in L_{ka} = \{\mathcal{T}_{ka}^1, \mathcal{T}_{ka}^2, \dots \}$. 
A pseudo-code for the teaching process is presented in Alg.~\ref{alg:teach}.

\begin{definition}
\label{def:teach}
A teaching tree $\mathcal{T}_{Teach}$ upon receiving a query as a state $s_{q}$ sequence in a message buffer $Q_m$, checks through a known states list in $L_{ks}$ and if present, responds through an appropriate state-action tree $\mathcal{T}_{ka}^*$, where $\mathcal{T}_{ka}^* \in L_{ka}$, a list of known action sub-trees.
\begin{align}
    \mathcal{T}_{Teach} &=\mathcal{T}_{ka}^*=Teach(Q_m) \nonumber \\ 
        &= \begin{cases} 
      \mathcal{T}_{ka}^i  & if \phantom{.}\exists\phantom{.} s_q \in Q_m \phantom{.} \mid \phantom {.} \hphantom{(} s_q = L_{ks}^i \in L_{ks}
      \\& \textit{, for some i} \phantom{.}and\phantom{.} \mathcal{T}_{ka}^i \in L_{ka}\\
      \emptyset & \textrm{otherwise (no response)}\\
   \end{cases}
   \label{eq:teach}
\end{align}
\end{definition}

\algnewcommand{\LeftComment}[1]{\Statex \(\triangleright\) #1}

\begin{algorithm}[t]
\caption{Pseudo code for \textbf{Teach} process.}
\label{alg:teach}
\begin{algorithmic}
\State \textbf{Input:} $Q_m\leftarrow$ List of queries received from other agents
\State \textbf{Data:} $L_{ks}, L_{ka}$
\State \textbf{Result:} {Transmit $\mathcal{T}_{ka}^*$ and return \textit{success} or return \textit{failure}}
\LeftComment{\textit{Check if a query is received.} }
\If {$Q_m.length() \neq 0$}
    \State $s_q \gets Q_m.pop()$
\LeftComment{\textit{Compare the query sequence against the sequences in known states list} }
    \For {$i \gets 1$ to $L_{ks}.length()$}
\LeftComment{\textit{If the query sequence is known, generate a response with the appropriate knowledge sub-tree} }
        \If{$s_q = L_{ks}^i$}
            \State $\mathcal{T}_{ka}^i \gets L^i_{ka}$
            \State{$\mathcal{T}_{ka}^* \gets \mathcal{T}_{ka}^i$}
            \State $Transfer(\mathcal{T}_{ka}^*)$
            \State \textbf{return} Success
        \EndIf
            
    \EndFor
\EndIf
\State \textbf{return} \textit{failure}
\end{algorithmic}
\end{algorithm}

\subsubsection{Learn}
A learning tree is defined as below.
A pseudo-code for the learning process is depicted in Alg.~\ref{alg:learn}.
\begin{definition}
\label{def:learn}
A learning tree $\mathcal{T}_{Learn}$ when faced with an unknown sequence $s_q$, broadcasts $s_q$ and waits for a response $\mathcal{T}^*_{ka}$. If received before a time out, it is combined using a $Sequence$ operation on the query conditions to form a knowledge sub-tree $\mathcal{T}_{k}$. This sub-tree is merged at the new knowledge $\mathcal{T}_{K}$ sub-tree segment in $\mathcal{T}_{Control}$. The query sequence $s_q$ and $\mathcal{T}_{ka}^*$ are added to $L_{ks}$ and $L_{ka}$ sets, respectively at the $i^{th}$ position in the new knowledge placeholder.
\begin{eqnarray}
\mathcal{T}_{Learn}=Learn(s_{q},Q_r) \label{eq:learn} \\
Add(\mathcal{T}_k^{i},\{s_q,\mathcal{T}_{ka}^*\}) \, , if \, \mathcal{T}_{ka}^* \neq \emptyset \label{eq:learn-update}
\end{eqnarray}
\end{definition}

\begin{algorithm}[ht]
\caption{Pseudocode for \textbf{Learn} process}
\label{alg:learn}
\begin{algorithmic}
\State \textbf{Input:} $s_q, Q_r \leftarrow$ List of action sub-trees received
\State \textbf{Data:} $L_{ks},L_{ka}, \mathcal{T}_{Control}$  
\State \textbf{Result:} {Merge received sub-tree $\mathcal{T}_k = Sequence(s_{q},\mathcal{T}_{ka}^*)$ with control sub-tree $\mathcal{T}_{Control}$ and return \textit{success} or \textit{failure}}
\LeftComment{\textit{If the current state sequence is not in the known-states list, then generate a query containing the current state sequence} }
        \If{$s_q$ \textbf{not} in $L_{ks}$}
            \State $Broadcast(s_q)$
        \EndIf
\LeftComment{\textit{Check if there is a response received.} }
\While {\textbf{not} timeout}
\If {$Q_r.length \neq 0$}
    \State $\mathcal{T}_{ka}^*=Q_r.pop()$
    \State $L_{ks}.add(s_q)$
    \State $L_{ka}.add(\mathcal{T}_{ka}^*)$
    \State $T_k\gets Merge(Sequence(s_q,\mathcal{T}_{ka}^*))$
    \State $\mathcal{T}_{Control} \gets Merge(\mathcal{T}_{Control},\mathcal{T}_k)$

    \State \textbf{return} \textit{success}
\EndIf
\EndWhile
\State \textbf{return} \textit{failure}
\end{algorithmic}
\end{algorithm}

\subsubsection{Timers} 
Finally, we define the two new timer decorators that are used in our KT-BT framework as follows. The pseudo-code versions of these two new timers are provided in Algorithms \ref{alg:tt1} and \ref{alg:tt2}.
\begin{definition}
\label{def:tt1}
\textit{A timer of type 1 (TT1), runs its child sub-tree $\mathcal{T}_{child}$ once after the time elapsed is greater than a set limit $t^1_{limit}$, returns a success after successfully running the child tree and a failure otherwise.}
\begin{equation}
    \mathcal{T}_{t1}= TT1(\mathcal{T}_{child},t^1_{limit})
    \label{eqn:tt1}
\end{equation}
\end{definition}

\begin{definition}
\label{def:tt2}
\textit{A timer of type 2 (TT2) runs its child sub-tree till the time elapsed is less than  $t^2_{limit}$, returns a success while running and a failure when stopped.}
\begin{equation}
    \mathcal{T}^2_{t2}=TT2(\mathcal{T}_{child},t^2_{limit})
\label{eqn:tt2}
\end{equation}
\end{definition}

\subsection{The StringBT representation of BT grammar}
\label{sec:stringbt}
The KT-BT framework requires a standard grammar for transmitting the response behavior tree $\mathcal{T}^*_{ka}$ by a teaching tree. Through this grammar, a sub-tree as a whole is transmitted through this grammar as a response to the queries posted by other robots. Therefore, we developed a unique \textit{stringBT} representation similar to the grammatical representation of behavior trees by Neupen et al. \cite{neupane2019learning} and Suddrey et al. \cite{Suddrey2021}.

In the \textit{stringBT} representation, all the generic BT operators are designed to have shorthand tags for their equivalent code formats in behavior tree constructs. 
 
The primary purpose of this grammatical representation is to simplify communication between agents and also to improve the human-readability aspect. For e.g., a sequence operator in \textit{stringBT} is represented as $<sq>$ followed by other operations. 
A summary of various \textit{stringBT} tags is presented in Sec.~\ref{sec:stringbt-sar} along with an implementation of this grammar.

In KT-BT, when a condition sequence is queried, a teaching robot responds with $\mathcal{T}^*_{ka}$ formatted as a \textit{stringBT}, and hence the response is the form of \textit{stringBT} sentences. On the receiving end, direct string manipulations like merge and append are performed using the received message at the \textit{stringBT} equivalent of $\mathcal{T}_{Control}$ (specifically, at the new knowledge placeholder part of the \textit{stringBT} grammar). The resultant $\mathcal{T}_{Control}$, which is also in \textit{stringBT} form is converted into generic code representations for (re-)compilation and ticking. 

The string constructs in the \textit{stringBT} grammar make it easier to search through its current BT during the teaching phase as well as merge operations during the learning phase. 
Furthermore, this gives the capability to generalize this structure across multiple domains and applications in robotics and MRS. 
Also, with advanced string manipulation techniques, it is also possible to relax the condition-action splitting requirement for every knowledge as assumed in Sec.~\ref{sec:teach}, as well as create the possibility of optimizing the BT and re-organizing the sub-trees (e.g., changing the priorities) in some applications.

\begin{algorithm}[t]
\caption{Pseudocode for Timer Type 1 \textbf{TT1}.}
\label{alg:tt1}
\begin{algorithmic}
\State \textbf{Input:} {$t^1_{limit}, \mathcal{T}_{child}$}
\State \textbf{Result:} {Ticks a sub-tree $\mathcal{T}_{child}$ once after a duration of $t^1_{limit}$ from the time the TT1 is ticked first.}
\LeftComment{\textit{On the first tick of the timer when it is set, store start time}} 
    \If{Timer.Start is \textit{True}}
        \State $StartTime \leftarrow Time.current()$
    \Else{}
        \State $TimeElapsed \leftarrow Time.current()-StartTime$
        \If{$TimeElapsed \geq t^1_{limit}$}
            \State $Run(\mathcal{T}_{child})$
            \State \textbf{return} Success
        \Else{}
            \State \textbf{return} Failure
        \EndIf    
    \EndIf
\end{algorithmic}
\end{algorithm}

\begin{algorithm}[t]
\caption{Pseudocode for Timer Type 2 \textbf{TT2}.}
\label{alg:tt2}
\begin{algorithmic}
\State \textbf{Input:} {$t^2_{limit}, \mathcal{T}_{child}$}
\State \textbf{Result:} {Ticks a sub-tree $\mathcal{T}_{child}$ for a duration of $t^2_{limit}$.}
\LeftComment{\textit{When the timer is set, store start time}} 
    \If{Timer.Start is \textit{True}}
        \State $StartTime \leftarrow Time.current()$
    \Else{}
        \State $TimeElapsed \leftarrow Time.current()-StartTime$
        \If{$TimeElapsed \leq t^2_{limit}$}
            \State $Run(\mathcal{T}_{child})$ \Comment{Tick Child}
            \State \textbf{return} Success
        \Else{}
            \State \textbf{return} Failure
        \EndIf    
    \EndIf
\end{algorithmic}
\end{algorithm}

\section{Formalization of the Knowledge Transfer}
\label{sec:formulation}
Here, the goal is to have knowledge shared between multiple agents involved in a mission. Having presented the definitions, we present more characteristics and technical analyses of the knowledge transfer process. We formalize the knowledge spread through the following lemmas.

First, we prove the knowledge transfer capability in KT-BTs (Lemma~\ref{lem:transfer}), followed by knowledge propagation (Theorem~\ref{lem:spread}) and the minimum opportunity requirement for maximum knowledge spread (Theorem~\ref{lem:opportunity}).

\begin{lemma}[\textbf{Knowledge transfer between two agents}]
\label{lem:transfer}
\textit{For an agent $i$, if there is an unknown state sequence $s_q$, that is known to an agent $k$. If the agents $i$ and $k$ can communicate, then the knowledge of agent $k$ for the state sequence $s_q$ is transferred to the agent $i$. }

\textit{i.e., $\exists$ a state-sequence $s_q$} s.t $s_q\notin L_{ks}(i)$, and $\exists$ at least one interactive agent $k$ at time $t\in (0,\infty]$, s.t $s_q \in L_{ks}(k)$. Then as the agent $i$ faces state-sequence $s_q$ at time $t$, $s_q \in L_{ks}(i)$, $\mathcal{T}_{ka}^*\in {L}_{ka}(i)$, and $\mathcal{T}_{ka}^*\subset\mathcal{T}_{Control}(i)$. Hence the agent $i$ is guaranteed to gain new knowledge to respond to the unknown state $s_q$ by using the KT-BT framework.
\end{lemma}
\begin{proof}
As agent \textit{i} faces conditions in state-sequence $s_q$, the $\mathcal{T}_{Learn}$ tree verifies the condition is not in agent \textit{i's}, $L_{ks}(i)$, and hence generates a query $s_q$. As, the query is received by agent the interactive agent \textit{k} in which, the $\mathcal{T}_{Teach}$ verifies in $k's$ known condition list $L_{ks}(k)$ and transmits the sub-tree $\mathcal{T}_{ka}^*$ in response according to Definition~\ref{def:teach}. Agent $i$, merges this tree with the $\mathcal{T}_{Control}$ tree and adds the condition $s_q$ to the list $L_{ks}$ and $\mathcal{T}_{ka}^*$ to ${L}_{ka}$.
\end{proof}

We expand the above lemma to all the agents in the group through the following theorem.

\begin{theorem}[\textbf{Knowledge spread across the entire group}]
\label{lem:spread}
\textit{In an MRS group of size $P$, if there is only one agent $k$ that has knowledge about a state sequence $s_q$, then the knowledge corresponding to $s_q$ is shared with all the agents in the group, as time $t \rightarrow \infty$.}

\textit{i.e., $\exists$ a one and only agent k that has the knowledge tree $\mathcal{T}_{ka}^{*}$ for a state-sequence $s_q$ belonging to a multi-agent group $G$ of population size $n(G)=P$. 
As time $t\rightarrow \infty$, }
\begin{center}
    $\forall G_j \in G, s_q \subset L_{ks} (j),\mathcal{T}_{ka}^{*} \in L_{ka}(j)$ and $\mathcal{T}_{ka}^{*} \subset \mathcal{T}_{Control}(j)$
    $\forall j= 1 \dots P$
\end{center}
\end{theorem}
\begin{proof}
Assume that all the agents in $P$ can interact with each other, and when an agent \textit{i} faces condition that is not in its known list of state sequence (knowledge database) $s_q \notin L_{ks} (i)$, also $s_q \in L_{ks} (k)$. By Lemma~\ref{lem:transfer}, the knowledge is transferred from agent $k$ to $i$, i.e., $s_q \in L_{ks} (i), \mathcal{T}_{ka}^*\in L_{ka} (i)$, and $\mathcal{T}_{ka}^*\subset\mathcal{T}_{Control}(i)$. This can also be proven to any agent \textit{j} within the communication range of \textit{i} or \textit{k}. 
Through this one-to-one transmission of knowledge after a sufficient amount of time, the sub-tree $\mathcal{T}_{ka}^{*}$ related to the state sequence $s_q$ is transferred to all the agents in the group $G$.
\end{proof}

We now define the lower bound of the number of occurrences of queries (opportunities) in the following theorem.

\begin{theorem}[\textbf{Opportunity of knowledge spread}]
\label{lem:opportunity}
In a group $G$ of size $n(G)=P$, if there is only one agent $k$ with knowledge of the state sequence $s_q$, then the minimum number of occurrences $N_{occ}$ (queries) of $s_q$ that are required for the knowledge $T_{ka}^*$ to be transferred to all the agents in the group is equal to $P-1$.

i.e. If $s_q \notin L_{ks} (j) \forall j \in [1,P]-\{k\}$, and $s_q \in L_{ks}(k)$, 
\centering {then $min\{N_{occ}(s_q)\} = P-1$} .
\end{theorem}
\begin{proof}
Assuming that each agent in the group $G$ faces the same unknown sequence $s_q$ only once, and all agents can interact with each other. Then a query is posted for every occurrence of $s_q$, corresponding to agents in $G$. The agents with the knowledge of $s_q$ address this query starting with agent $k$, as this is the only agent with the knowledge of the sequence $s_q$ initially (according to lemma 2). In this process, the total number of queries posted is $P-1$ (queries by all agents except agent $k$). 
If the occurrences of $s_q < P-1$, then there will be some agents that will not have faced the state sequence $s_q$ and hence will never gain its corresponding knowledge. 

Therefore, at least $P-1$ queries of the same knowledge $s_q$ would be needed to guarantee propagation of that knowledge to the entire group, as long the queries do not come at the same time, and at least one of the agents in the group has that knowledge in its knowledge base ($\mathcal{T}_{k}$). 
In other words, $N_{occ}(s_q)$ is nothing but the opportunities provided to the agents in $G$ to learn the knowledge corresponding to state $s_q$ from each other. 
\end{proof}

The actual number of queries (or opportunities) would depend on the connectivity graph, the number of neighboring agents that can respond to the query, the response rate, and the need to require the knowledge with $s_q$ in the mission. For instance, if agent $k$ is at the center of the connectivity graph, the knowledge spread will be faster than this agent being at the end of a line graph, for example. In addition, the opportunity for propagation will be higher (fewer queries) if more than one agent has the same knowledge that can be shared.

\begin{figure}[t]
    \centering
    \includegraphics[width=0.9\linewidth]{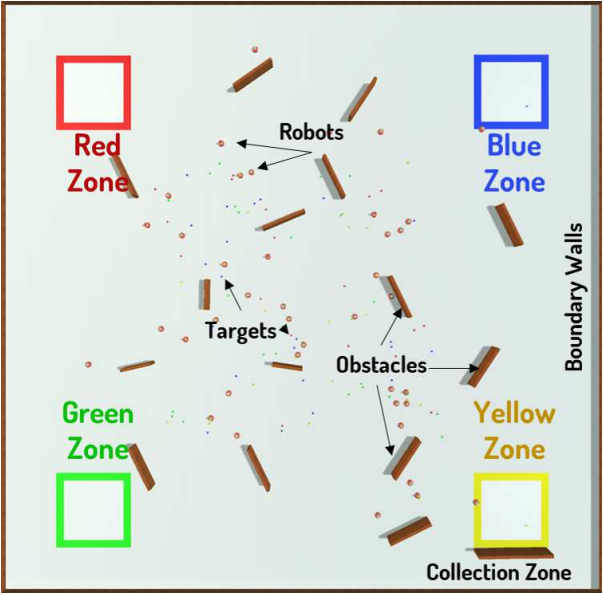}
    \caption{Sample search space showing four collection zones at corners, randomly located targets, and multiple robots performing SAR simulation.}
    \label{fig:SearchSpace}
\end{figure}

\section{Case Study Application: Search and Rescue}
\label{sec:SAR}
To provide an example of the concepts defined earlier and to analyze the framework, we consider a Search and Rescue (SAR) problem with multiple robots. The SAR problem aims to collect different color-coded targets and move them to their corresponding collection zones. 
The generalized SAR problem we used here is analogous to multi-robot foraging and multi-target search problems. These problems are predominantly used to test multi-robot algorithms \cite{shell2006foraging,harwell2018broadening}.

\subsection{Search Space}
The search space is a rectangular space defined by $\mathcal{A}=[0,x] \times [0,y]$ dimensions.
The targets are cubes colored in red, green, yellow, and blue. There are four collection zones for each of the colored targets located at the four corners of the configuration space $\mathcal{A}$. The number of red, green, yellow and blue targets are $n_r,n_g,n_y$ and $n_b$ respectively and the total number of targets $n_t=n_r+n_g+n_y+n_b$. The targets are randomly scattered on the 2D plane $\mathcal{A}$, and both the targets and collection points are stationary. The configuration space may or may not have obstacles; however, every robot perceives other robots as obstacles. Fig.~\ref{fig:SearchSpace} presents a sample search space with randomly distributed targets.

\begin{figure}[t]
    \centering
    \includegraphics[width=0.49\textwidth]{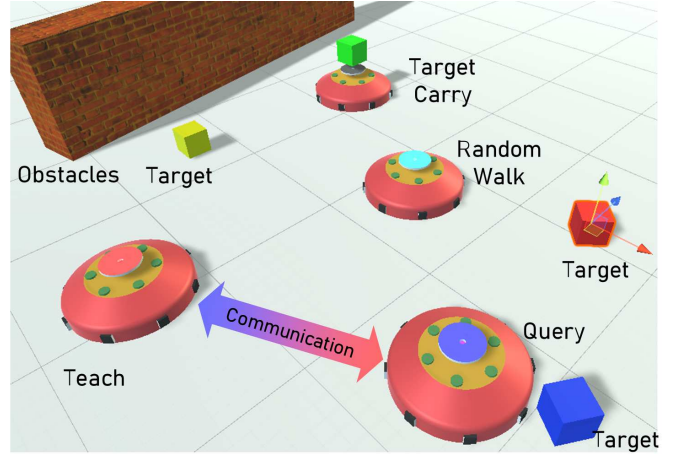}
    \caption{A snapshot of simulation, showing various robot states. A robot indicator on the top blinks in red when teaching, blue during a query, and cyan during the random walk.}
    \label{fig:RobotStates}
\end{figure}

\subsection{KT-BT SAR Simulator}
To test the KT-BT framework, we developed a simulator tool for the SAR problem in the Unity 3D game development environment (see Fig.~\ref{fig:SearchSpace}). We used the Fluid BT library\footnote{\url{https://github.com/ashblue/fluid-behavior-tree}} and adapted them for the  KT-BT framework by combining them with the Roslyn\footnote{\url{https://github.com/dotnet/roslyn}} framework. 
We specify a code segment of BT called \textit{LiveBT}, which is the compiled version of the knowledge base $\mathcal{T}_{control}$, and this \textit{LiveBT} controls the robot based on its status. 
In general, the Fluid BT libraries are designed to have the trees pre-compiled before the start of simulations like any other Behavior Tree library for robotics\footnote{\url{https://www.behaviortree.dev/}}. 
However, in our KT-BT framework, the BT needs to be updated in real time while the BT is being used for robot control. 

Specifically, the \textit{LiveBT} should be re-compiled every time a change is made in the form of new knowledge incorporated (through learning) without affecting its current execution. 
Therefore, we utilized the Roslyn framework's real-time compilation capabilities to address this challenge. Here, we use the \textit{stringBT} version to recompile $\mathcal{T}_{Control}$ and update the \textit{LiveBT}. For every change, a compiled BT is stored back in \textit{LiveBT} and is ticked immediately.

The robots in the simulator indicate their states through the colored lights on the top, as shown in Fig. \ref{fig:RobotStates}. For example, a robot in query mode blinks blue light, and a robot in teach mode blinks red. A complete simulator with interactive GUIs for testing all the simulation modes and strategies is available\footnote{\url{https://github.com/herolab-uga/KTBT-Release} We also provide support documentation in the link for running the simulations. The simulator can run multiple instances in parallel and scale to hundreds of robots depending on the available hardware resources. The readers can obtain additional experimental data with this simulator if needed.}. 

\begin{figure*}[ht]
    \centering
    \includegraphics[width=0.98\textwidth]{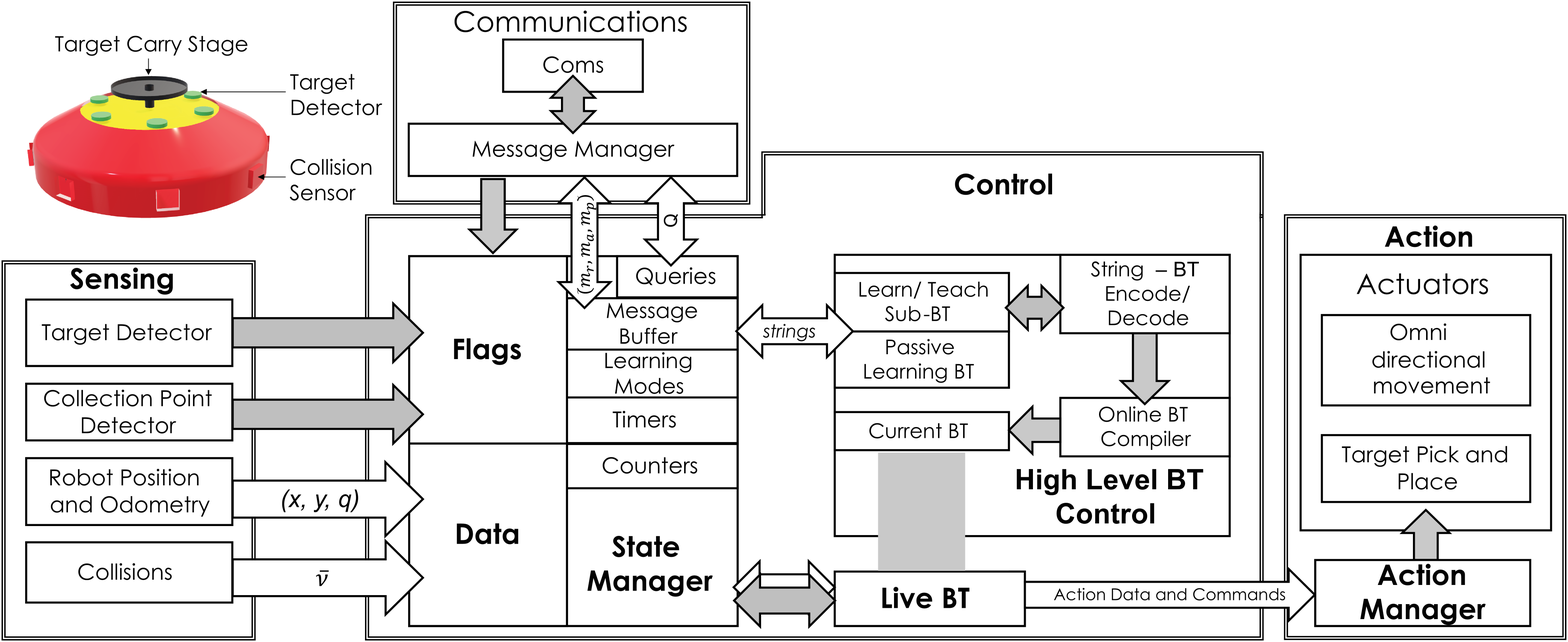}
    \caption{3D Model of a Robot showing collision detectors, target detectors, and a target carry stage (inset). Robot's functional architecture shows various modules and connections between them. These modules can be broadly segregated under sensing, communication, control, and action. 
    The picture inset shows the robots interacting in the SAR simulator.}
    \label{fig:RobotArchModel}
\end{figure*}

\subsection{Robot Model Architecture}

The agents in this study are mobile robots that are equipped with sensors for target detection in close range, collection zone detection, and collision detection. They also have actuation mechanisms for target pickup and omnidirectional movement on a 2D plane. In addition to these, the robots are equipped with a communications module for broadcasting queries and receiving responses. The robot decision-making is carried out in a controller module that generates control actions from condition sequences, called \textit{LiveBT} controller. The final control information is sent to the actuators to perform live actions like target pickup and movement. A complete robot model architecture and a physical model of the robot used in the simulator are presented in Fig. \ref{fig:RobotArchModel}.

\subsubsection{Sensing}
The sensing module has a suite of four different sensors for target detection, collection zone detection, robot position and odometry, and collision detection. A target detector detects the presence of a target in its range $D_T$ along with the target type (R, G, Y, or B). A collection point detector detects if the robot is entirely inside a collection zone along with the target type it is carrying. The robot also gets its position with respect to a global coordinate system through its position, and odometry sensor suite in a tuple $\langle Position, Orientation \rangle$, where $Position$ is a position vector and $Orientation$ is a quaternion. 

Finally, a collision detector detects all possible collisions with adjacent robots and other objects in the configuration space falling within a field collision of range $D_c$ defined by 
\begin{equation}
    F_{collision}= \begin{cases}
      0 & :D_{obj}>D_c\\
      1 & :D_{obj}\leq D_c\\
    \end{cases}  
\end{equation}

$\forall$ the points of collision $C$ in detection range $D_c$, a resultant vector is computed as
\begin{equation}
    V_{collision_i}=\begin{cases}
        C_i-P &: F_{collision_i} = 1\\
        0     &: F_{collision_i} = 0
    \end{cases}
\end{equation}
where $i = 1 \dots n(C)$ and subscripts are the indices of the $i^{th}$ object.

Finally, a resultant vector for all the collision vectors is computed as 
\begin{equation}
    V_c=\sum_{i = 1}^{n(C)}V_{collision_i}
    \label{Eq:ColAvoid}
\end{equation}

\subsubsection{Communications}
A communications module establishes generic communication channels between robots falling within a range $D_{coms}$. These channels can broadcast and receive messages, generally composed of queries, responses, and corresponding flags. E.g., a learn tree sends a query to the message manager through the state manager to broadcast, and a response received through the same path is handled by the learn tree accordingly. The behavior trees are encoded in \textit{stringBTs} during transfer.

\subsubsection{Actions}
The actions module contains an action manager, which translates the controller output signals to actions in the environment. Every robot has an omnidirectional movement actuator and a target pick-place actuation mechanism. The general commands to the action manager include the direction of movement, speed, angle of rotation, pick and place. A target picked up is carried on the target carry stage on the top of the robot, as shown in Fig.~\ref{fig:RobotStates}.

\subsubsection{Control}
The important decisions of robot planning, learning, and teaching are made in the Control module through BTs. A control module is divided into three sub-modules, a State Manager, High-Level BT Control, and a Live BT module. A state manager, similar to a blackboard, maintains the status of various internal and external flags and conditions. The external flags correspond to the state of the sensors and communicators, and internal flags, on the other hand, represent the robot states for seamless decision-making at the BT level. In addition to these flags, State Managers also handle data from the sensors like collision vectors, robot position, and odometry, communication queries, and manage message buffers, counters, and timers.

A High-Level BT Control manages the core behavior tree $\mathcal{T}_{kt}$ according to the BT presented in Definition~\ref{def:kt}. The $\mathcal{T}_{Control}$ portion of the main BT  in High-Level BT Control is stored as a \textit{stringBT}, and the corresponding actual $\mathcal{T}_{Control}$ is a compiled version of the represented \textit{stringBT} which is ticked at regular intervals. The controller module is designed to compile the \textit{stringBT} whenever a change is detected in the \textit{stringBT} version of $\mathcal{T}_{Control}$ of High-Level BT Control and stored as \textit{LiveBT}. The following sub-sections present further details on the BT design, a \textit{stringBT} encoding example, and the challenges of real-time compilation.

\begin{figure*}[t]
    \centering
    \includegraphics[width=\textwidth]{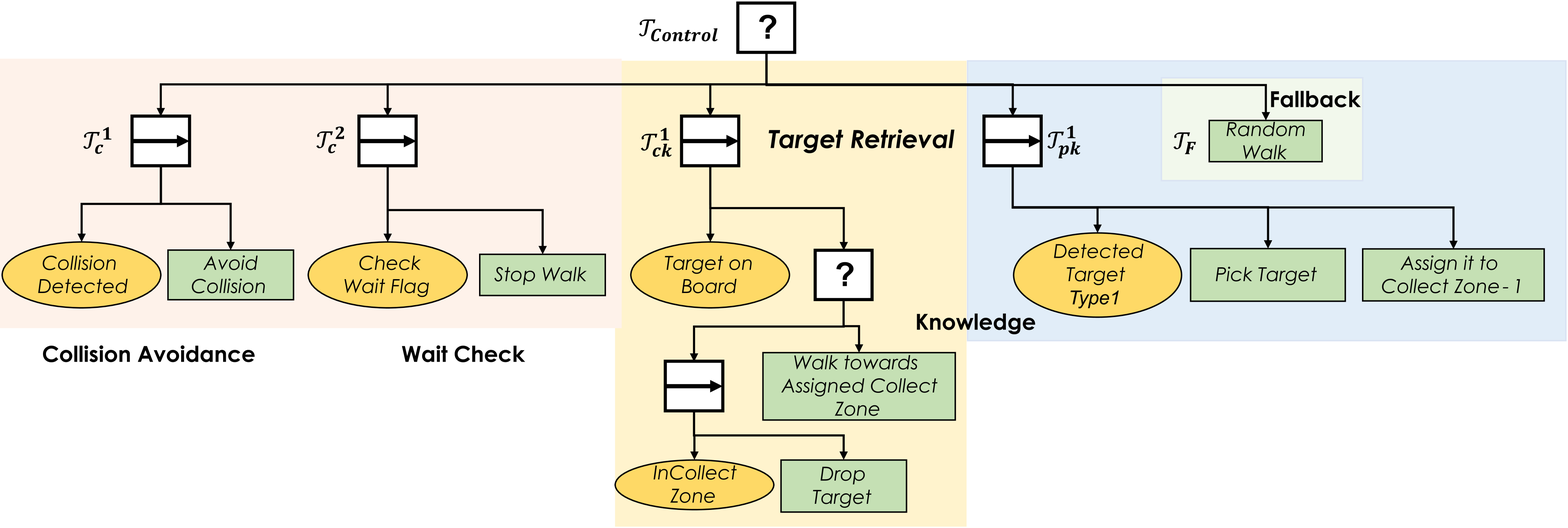}
    \caption{Example control tree used on the robots for SAR simulation case study.}
    \label{fig:ControlTreeEx}
\end{figure*}

\subsection{Behavior Tree design}
The High-Level BT Control sub-module in the controller maintains a behavior tree of a structure following Definition~\ref{def:kt}, i.e., the trees $\mathcal{T}_{Control}$, $\mathcal{T}_{Teach}$ and $\mathcal{T}_{Learn}$ running in parallel. 
\subsubsection{Control Sub-Tree}
According to Definition~\ref{def:control}, a control sub-tree $\mathcal{T}_{Control}$ should contain a selector with sub-trees in the order of criticality followed by action, knowledge, and fallback sub-trees. The current robot models are designed to have two critical sub-trees. The first critical sub-tree $\mathcal{T}_{c}^1$ is designed for collision avoidance followed by the second wait sub-tree $\mathcal{T}_{c}^2$, as shown in the Fig.  \ref{fig:ControlTreeEx}.

In the collision avoidance sub-tree, when a collision flag is \textit{true} in the state manager, its corresponding mean collision vector $V_c$ is computed from Eq.~\eqref{Eq:ColAvoid}. A unit vector in the direction $-V_c$ is computed in the AvoidCollision action and the corresponding control command is sent to the action manager.

The next sub-tree following the critical sub-trees sequence is the common knowledge sub-tree sequence $\mathcal{T}_{CK}$. In the current SAR problem, this is a target retrieval sub-tree, which is common across all the robots. This sub-tree ensures that if any target is picked up or on board, it is moved to its assigned collection zone. This action sub-tree $\mathcal{T}_{ck}^1$ is shown in Fig.  \ref{fig:ControlTreeEx}.

The sub-trees following the common knowledge sub-tree sequence are for the prior knowledge $\mathcal{T}_{PK}$. This is a placeholder location for the prior knowledge sub-trees. For example, the knowledge sub-tree shown in Fig. \ref{fig:ControlTreeEx}, is for retrieving target type 1 (R-Red). As this sub-tree is already part of the control sequence, its condition sequence is also a sub-set of $s_{ka}$, and the robot, when queried, can respond with the $\mathcal{T}_{pk}^1$ as $\mathcal{T}_{ka}^*$.

Finally, the sub-tree to the extreme right is a fallback tree $\mathcal{T}_F$, which is executed when none of the sub-trees to the left return a success. In the SAR case, it is the random walk action, where the robot chooses a random direction and walks for a certain duration.

\subsubsection{Teach Tree}
\begin{figure}[t]
    \centering
    \includegraphics[width=\linewidth]{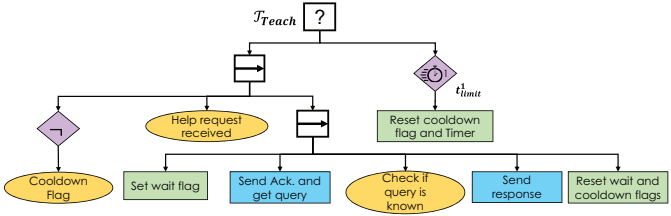}
    \caption{The teaching behavior tree in SAR robots.}
    \label{fig:TeachTreeEx}
    \vspace{-10pt}
\end{figure}

The Teach sub-tree is similarly structured in all robots. From Definition~\ref{def:teach}, a Teach tree continuously checks for any queries being broadcast and further checks if the query is in its condition set. If found, it responds with the appropriate knowledge subtree, and additionally, a cool-down flag is checked every time a query is encountered to ensure the robot is not stuck in a Teach loop when multiple robots are querying simultaneously. This cool-down flag is reset after a time $t^1_{limit}$, run by a TT1 decorator, as shown in Fig. \ref{fig:TeachTreeEx}.

\subsubsection{Learn Tree}

\begin{figure}[t]
    \centering
    \includegraphics[width=0.9\linewidth]{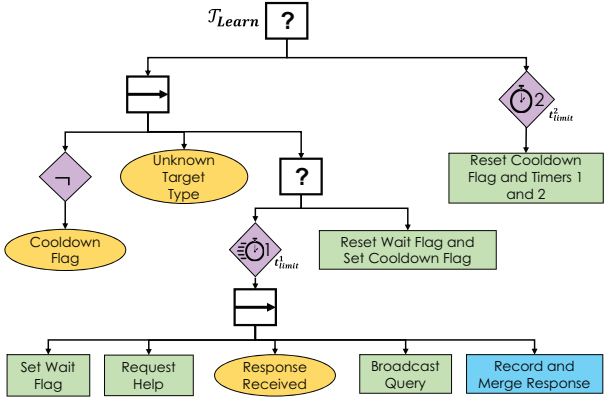}
    \caption{The learning behavior tree in SAR robots.}
    \label{fig:LearnTreeRLEx}
\end{figure}

A Learn tree $\mathcal{T}_{teach}$ runs in parallel to the control and teaching trees. The Learn tree designed for the current SAR problem is shown in Fig. \ref{fig:LearnTreeRLEx}.

The designed learning tree, $\mathcal{T}_{Learning}$ from Definition~\ref{def:learn}, has two timers which are decorator nodes with two different functionalities, as shown in Fig.~\ref{fig:LearnTreeRLEx}. Timer type 1, is a pulse timer defined previously in Definition~\ref{def:tt1}, and timer type 2 is a run timer as per Definition~\ref{def:tt2}. In contrast to timer type 1, the type 2 timer runs the associated behavior tree as long as the timer is running.

The type 2 timer serves the purpose of query and wait, where the robot queries about an unknown target and waits for a duration $t^2_{limit}$. The BT, in this case, is designed to check the cool-down flag before executing the query sequence. This is a common flag shared between the Teach BT in Fig. \ref{fig:TeachTreeEx}, and hence a cool-down flag set will run the Teach and Learn trees in the wait loops controlled by TT1 timers. This is to avoid repeated detection and queries on the same target when no response is received.

\subsection{StringBT implementation of SAR application}
\label{sec:stringbt-sar}
In our proposed grammar, we assume that the sets of conditions flags and actions are appropriately labeled in both state and action managers. For example, an action stating \textit{RandomWalk} is an action routine that can be initiated with the tag \textit{'RandomWalk'}.
While designing the grammar for the BT representation, the rules were written for BT encoding inline with the \textit{FluidBT} library in Unity, and wrappers for these rules were written to convert the \textit{stringBT} structures to the \textit{FluidBT} codes. We summarize some of the grammar rules formulated for \textit{stringBT} representation and \textit{FluidBT} equivalent in Table \ref {table:stringbt-sar}. An example BT representation of \textit{stringBT} and \textit{FluidBT} code is presented in Alg.~\ref{alg:stringbt}.

\begin{algorithm}[t]
\caption{\textbf{stringBT for $\mathcal{T}_{Control}$ in the SAR case study.}}
\label{alg:stringbt}
\begin{algorithmic}
\State \hspace{-1em} $<$Root$>$\\
\hspace{0em}$<$sl$>$\\
\hspace{0.5em}$<$sq$><$c$>$(\_collisionDetectedF)\\
\hspace{3em}$<$a$>$(CollisionAvoidance)$<$e$>$\\
\hspace{0.5em}$<$sq$><$c$>$(\_waitF)\\
\hspace{3em}$<$a$>$(StopWalk)$<$e$>$\\
\hspace{0.5em}$<$sq$><$c$>$(\_treasureOnBoardF)\\
\hspace{3em}$<$sl$><$sq$><$c$>$(\_inZoneF)\\
\hspace{7.75em}$<$a$>$(PlaceTreasure)$<$e$>$\\
\hspace{5.25em}$<$sq$><$c$>$(!\_inZoneF)\\
\hspace{7.75em}$<$a$>$(WalkToCollection)$<$e$><$e$><$e$>$\\
\hspace{0.5em}$<$a$>$(RandomWalk)$<$e$>$
\end{algorithmic}
\end{algorithm}

\begin{table}[ht]
\caption{BT Operator equivalents in \textit{FluidBT} and \textit{stringBT} frameworks.}
\label{table:stringbt-sar}
\centering
\begin{tabular}{lll}
\toprule
\textbf{BT Operator}	& \textbf{\textit{FluidBT}} &\textbf{stringBT}\\
\midrule
Sequence                             &.Sequence() &$<sq>$\\
Selector                             &.Selector() &$<sl>$\\
Parallel                             &.Parallel() &$<pl>$\\
Condition                            &.Condition($()\Rightarrow$ State) &$<c>$\textit{ConditionTag}\\
Action                   &.Action($()\Rightarrow$Method, $r$) &$<a>$\textit{ActionTag}\\
Wait                     &.Wait(WaitDuration) &$<w>$\textit{WaitDuration}\\
Segmentation            &.End() &$<e>$\\
\bottomrule
\end{tabular}
\end{table}

\subsection{Implications on Real-World Robot Implementations}
\label{sec:realrobot}

In this paper, we chose to validate the framework in a simulation environment since there are a few technical challenges to deploying the KT-BT framework on real-world robots. Most real-world robots use Linux-based Robot Operating Systems (ROS\footnote{\url{https://www.ros.org/}}) as their software framework. ROS-compatible software tools\footnote{\url{https://github.com/BehaviorTree/Groot}}\footnote{\url{https://github.com/BehaviorTree/BehaviorTree.CPP}} available currently for the design and visualization of BTs support only pre-compiled BT structures. This limitation does not allow dynamic (real-time) updates or re-compilation of BTs for knowledge updates while the BT is being used by the robot for execution. We plan to overcome this challenge by developing wrappers similar to the Roslyn framework compatible with ROS and implementing them on a swarm robotics test bed.
In principle, the KT-BT framework is feasible for real-world robots by addressing the above technical challenges.

\begin{table}[t]
\caption{A summary of various types of studies conducted.}
\label{table:summary}
\begin{tabular}{ll}
\toprule
\textbf{Type of Study}	& \textbf{Goals} \\
\midrule
No Transfer Vs. KT-BT & \multicolumn{1}{m{5cm}}{Compare performance of KT-BT and No Transfer, and for configurations with and without obstacles.}\\
Opportunities &\multicolumn{1}{m{5cm}}{Study the effect of opportunities on knowledge spread and performance.} \\
Communication Range &\multicolumn{1}{m{5cm}}{Study the effect of communication range on group performance, knowledge spread, and query efforts} \\

\bottomrule
\end{tabular}
\end{table}

\section{Experimental Analysis}
\label{sec:experiment}
On the KT-BT SAR simulator, we conducted studies to understand the group performance, opportunities, knowledge spread, query efforts, the effect of communication range, opportunities, and heterogeneity trends in various scenarios. These studies are summarized in Table~\ref{table:summary}.

Across these studies, we maintain six different types of robots based on their prior knowledge levels. These are labeled as Ignorant $(I)$, Multi-target $(M)$, Target-Red $(R)$, Target-Green $(G)$, Target-Yellow $(Y)$, and Target-Blue $(B)$. An ignorant robot has no prior knowledge of handling any target type. And on the other hand, a Multi-target robot can handle any target type. The rest of the robot types have prior knowledge about the color they are associated with. For the current study, we use different combinations of these robots to evaluate the groups' performance. For example, a combination of $(10,10,5,5,5,5)$ has robots of numbers in the sequence $(I,M,R,G,Y,B)$. To maintain sufficient space for movement and avoid crowding, we kept the total number of robots at 40 across all our studies.

\subsection{No Transfer Vs. KT-BT study}

In this study, we compare the performance of three different groups that differed in their knowledge transfer capabilities, as shown in Table \ref{Table: NTvsKTBT}. Base Line 1 (BL1) group consists of agents with knowledge of handling any target type, and Base Line 2 (BL2) has agents that are evenly grouped to drive each target type. In BL2, agents cannot transfer knowledge; otherwise, the agents cannot query other agents for help with unknown conditions (Queries in BL1 do not arise as all the agents have complete knowledge). We compare the performance of these baseline groups with a KT-BT group that contained agent composition similar to BL2 and additionally is enabled with the knowledge transfer ability. Additionally, simulation trials were conducted in two different configuration spaces that varied in the presence of obstacles, as shown in Fig.~\ref{fig:SearchSpace}. In both configuration spaces, each target type was fixed at 25, and the position of the targets was randomly varied across all the trials. 

\begin{table}[t]
\centering
\caption{Simulation parameters for KT-BT study.}
\label{Table: NTvsKTBT}
\begin{tabular}{r|lllc}
\hline
\textbf{Parameter}                                                               & \multicolumn{4}{c}{\textbf{Value}}                                                                                                 \\ \hline
Sim Mode                                                                         & \multicolumn{3}{c|}{NT}                                                                                   & KT-BT                  \\ \cline{2-5} 
\begin{tabular}[c]{@{}r@{}}Robots Combination \\ (I, M, R, G, Y, B)\end{tabular} & \multicolumn{3}{c|}{\begin{tabular}[c]{@{}c@{}}(0, 40, 0, 0, 0, 0)\\ (0, 0, 10, 10, 10, 10)\end{tabular}} & (0, 0, 10, 10, 10, 10) \\ \cline{2-5} 
\begin{tabular}[c]{@{}r@{}}Target Combination \\ ( R, G, Y, B)\end{tabular}      & \multicolumn{1}{c}{}                            & \multicolumn{3}{c}{$(n_r, n_g, n_y, n_b) = (25, 25, 25, 25)$}                    \\\hdashline
Obstacles                                                                        &                                                 & \multicolumn{3}{l}{with and without}                                             \\
Communication Range                                                              &                                                 & \multicolumn{3}{l}{200 units}                                                    \\
Iterations                                                                       &                                                 & \multicolumn{3}{l}{50000}                                                        \\
Trials                                                                           &                                                 & \multicolumn{3}{l}{20}                                                           \\ \hline
\end{tabular}

\end{table}

\begin{figure}[t]
    \centering
    \includegraphics[width=0.98\linewidth]{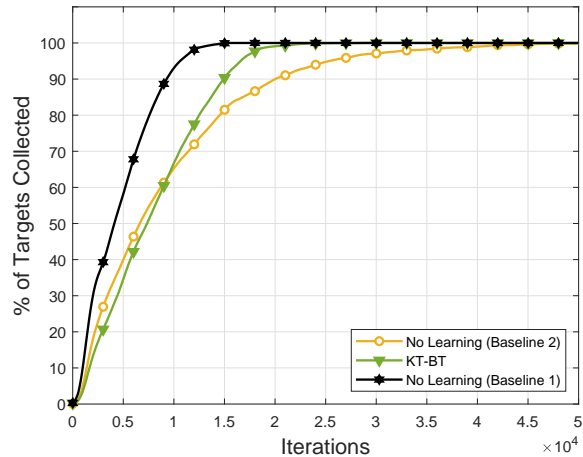}
    \caption{Average percentage of targets collected over time over 20 trials, for No-transfer with $(0, 40, 0, 0, 0, 0)$ (BL1), No-transfer with $(0, 0, 10, 10, 10, 10)$ (BL2) and KT-BT with $(0, 0, 10, 10, 10, 10)$ group compositions.}
    \label{fig:TS-NTvsKTBT}
\end{figure}

\begin{figure}[t]
    \centering
    \includegraphics[width=0.98\linewidth]{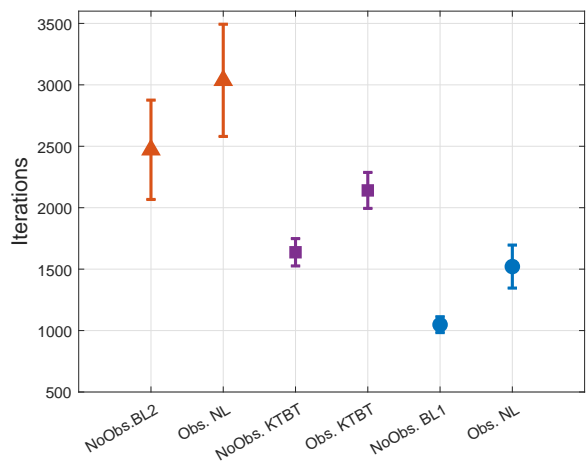}
    \caption{Efficiency of collecting $99\%$ of targets averaged over 20 trials for scenarios with and without obstacles.}
    \label{fig:ObsVsNoObs}
\end{figure}

The time series graphs for the total percentage of targets collected are presented in Fig.~\ref{fig:TS-NTvsKTBT}, and the performance comparison is made in Fig.~\ref{fig:ObsVsNoObs}. 
The percentage of target collection, shown in Fig.~\ref{fig:TS-NTvsKTBT}, is the average across 20 trials conducted for the same robots and target compositions, but the initial positions of the targets and robots were randomly varied. It can be noted that the worst performer was the baseline 2 (BL2) group, which lacked any knowledge transfer capabilities. Accordingly, the best performers were the baseline 1 (BL1) groups that had knowledge about all the target types. The true advantage of knowledge transfer can be noticed in the performance of the KT-BT groups that are similar in composition to BL2 groups but also could query and respond. From the time series graph Fig.~\ref{fig:TS-NTvsKTBT}, it can be noted that the KT-BT groups lagged BL2 groups initially, as the query-response process in the robots introduced delays. But, going further, the KT-BT group's performance surpassed BL2 as more robots learned to deal with multiple target types.

The mean performance graph over 20 trials measuring the number of iterations the groups took to collect $99\%$ of the targets is presented in Fig.~\ref{fig:ObsVsNoObs}. The graph also shows the collection rate decreased (higher number of iterations) in with-obstacle scenarios across all the groups but followed a similar trend as the no-obstacle scenario. 

\subsection{Knowledge Spread and Opportunities Study}
This study aims to understand the effect of opportunities on the knowledge spread in robots. Here the opportunities are the number of targets available in the configuration space. In this study, we varied the target counts between 10 and 100 of each color type, as shown in Table~\ref{Table: KnowOpp-Study}. We varied the target counts in the simulations that contained a single group with one multi-target and 39 ignorant robots (called 39I-1M group). 

Whenever a target is encountered, an ignorant robot posts a query with its adjacent neighbors and awaits a response. In this case, at the beginning of the simulations, only one robot can respond to any query. As the simulation progresses, the knowledge about various targets is shared among the groups; thus, the robots learn to handle multiple targets. 

\begin{table}[t]
\centering
\caption{Simulation summary for knowledge spread and opportunities study}
\label{Table: KnowOpp-Study}
\begin{tabular}{r|llll}
\hline
\textbf{Parameter}                                                          & \multicolumn{4}{c}{\textbf{Value}}                                                                                                          \\ \hline
Sim Mode                                                                    & \multicolumn{4}{c}{KT-BT}                                                                                                                   \\
\begin{tabular}[c]{@{}r@{}}Robots Combination \\ (I,M,R,G,Y,B)\end{tabular} & \multicolumn{4}{c}{(39,1,0,0,0,0)}                                                                                                          \\
\begin{tabular}[c]{@{}r@{}}Target Combination \\ ( R,G,Y,B)\end{tabular}    & \multicolumn{4}{c}{\begin{tabular}[c]{@{}c@{}}(0,0,10,10,10,10), (0,0,25,25,25,25)\\ (0,0,50,50,50,50), (0,0,100,100,100,100)\end{tabular}} \\ \hdashline
Obstacles                                                                   &                                    & \multicolumn{3}{l}{without}                                                                            \\
Communication Range                                                         &                                    & \multicolumn{3}{l}{200 units}                                                                          \\
Iterations                                                                  &                                    & \multicolumn{3}{l}{50000}                                                                              \\
Trials                                                                      &                                    & \multicolumn{3}{l}{20}                                                                                 \\ \hline
\end{tabular}
\end{table}

\begin{figure}[t]
    \centering
    \fontfamily{phv}\selectfont
    \includegraphics[width=0.98\linewidth]{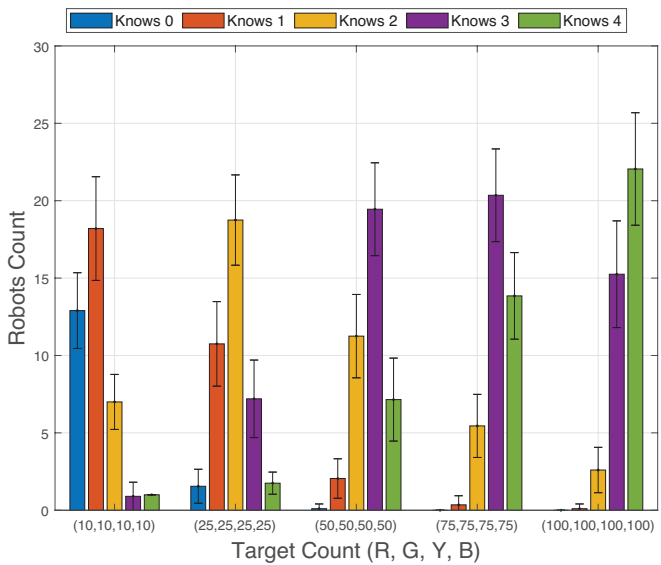}
    \caption{Graphs showing the counts of robots at various knowledge levels and at the end of a simulation (39I-1M).}
    \label{fig:RCount-Opportunity}
\end{figure}

\vspace{10pt}
We segregate robots into different levels based on the number of targets they can handle. For e.g., a robot that knows to handle two types of targets is grouped under ``knows – 2''; similarly, a robot that knows to handle all targets is grouped under ``knows – 4''. A robot starting in a ``knows – 0'' group progresses to higher level groups as more knowledge is acquired. In each trial, the final number of robots in all four groups is counted for different target counts (opportunity counts). The results of this experiment averaged over 20 trials are presented in Fig.~\ref{fig:RCount-Opportunity}. 

With the increase in the number of targets, more robots had the opportunity to gain knowledge about multiple targets, i.e., the count of robots with the knowledge to handle all four targets monotonously increased with the increase in the number of target opportunities, as seen in Fig.~\ref{fig:RCount-Opportunity}. The rise and drop in the counts of robots that know 3, 2, and 1 target types are because of the shift in numbers across groups when more opportunities were made available. 

From Fig.~\ref{fig:RCount-Opportunity}, it can be observed that the KT-BT framework was efficient in demonstrating knowledge transfer and spread in a multiagent system. Further, it can also be inferred from the graph that if each robot in a group size of $p$ is allowed to query and learn from only one target, for all the robots to gain complete knowledge, they require $p-1$ opportunities of each target type, thus validating the theorem \ref{lem:opportunity}.

\subsection{Effect of Communication Range on Knowledge Transfer}

\begin{figure}[t]
    \centering
    \includegraphics[width=0.98\linewidth]{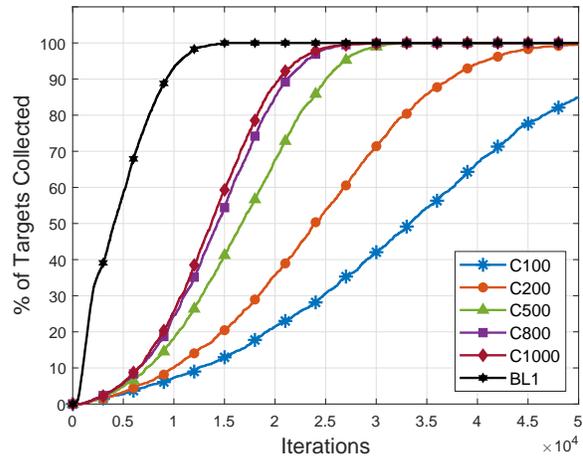}
    \caption{Target collection rate for different communication ranges in a group with 39 Ignorant and 1 Multi-target robot.}
    \label{fig:ComTarget}
\end{figure}

In this analysis, we varied the communication range of robots from 100 units to 1000 units. We maintained the population constant with 39 Ignorant and 1 Multi-target robot, as summarized in Table~\ref{Table: Com-Study}. We compared the target retrieval performance with BL1 and cumulative counts of lost queries for all the different communication ranges. In target retrieval performance comparison (see Fig.~\ref{fig:ComTarget}), the performance of the test group progressively improved with the increase in communication range, approaching the ideal BL1 performance.

\begin{figure}[t]
    \centering
    \includegraphics[width=0.98\linewidth]{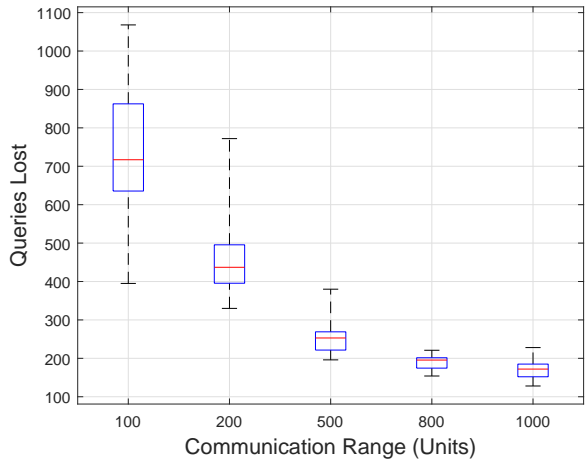}
    \caption{Effect of communication range on the losses in the number of queries.}
    \label{fig:ComEffect}
\end{figure}

From our study comparison of effective communication plotted from the query loss graph in Fig.~\ref{fig:ComEffect}, it is inferred that lower communication ranges resulted in higher query losses than larger communication ranges. This suggests the need for reliable and long-range communication for better knowledge transfer and, eventually, better group performance.

\begin{table}[t]
\centering
\caption{Simulation summary for the study on the effect of communication range on knowledge transfer}
\label{Table: Com-Study}
\begin{tabular}{r|llll}
\hline
\textbf{Parameter}                                                          & \multicolumn{4}{c}{\textbf{Value}}                      \\ \hline
Sim Mode                                                                    & \multicolumn{4}{c}{KT-BT}                               \\
\begin{tabular}[c]{@{}r@{}}Robots Combination \\ (I,M,R,G,Y,B)\end{tabular} & \multicolumn{4}{c}{(39,1,0,0,0,0)}                      \\
\begin{tabular}[c]{@{}r@{}}Target Combination \\ ( R,G,Y,B)\end{tabular}    & \multicolumn{4}{c}{(0,0,25,25,25,25)}                   \\
Obstacles                                                                   &          & \multicolumn{3}{l}{without}                  \\
\begin{tabular}[c]{@{}r@{}}Communication Range\\ (units)\end{tabular}       &          & \multicolumn{3}{l}{100, 200, 500, 800, 1000} \\
Iterations                                                                  &          & \multicolumn{3}{l}{50000}                    \\
Trials                                                                      &          & \multicolumn{3}{l}{20}                       \\ \hline
\end{tabular}
\end{table}

\begin{figure}[t]
    \centering
    \includegraphics[width=0.98\linewidth]{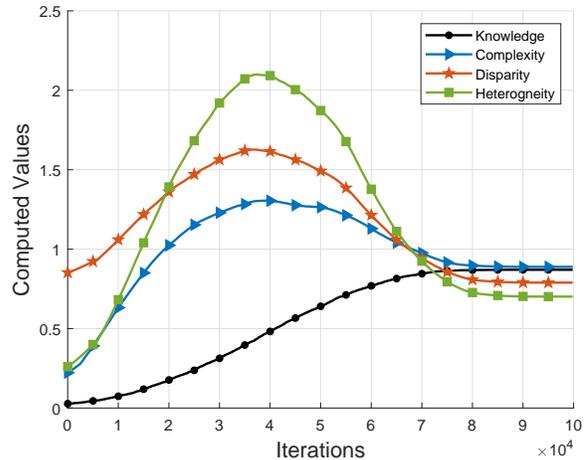}
    \caption{Complexity, Disparity, Heterogeneity and Knowledge score changes over iterations for 39 Ignorant and 1 Multi-target robot group (39I-1M) searching for a target combination $(R,G,Y,B)=(100,100,100,100)$}
    \label{fig:Hs391}
\end{figure}

\subsection{Knowledge and Functional Heterogeneity}

We extend the knowledge propagation study to estimate the functional heterogeneity in the group. Heterogeneity is measured as a product of complexity and disparity, as proposed by Twu et al. \cite{twu2014measure}, where complexity estimates how distributed the group is in its knowledge, and disparity measures how distinct these group members are in their knowledge. Complexity is computed as entropy for the distribution of agents across different species as shown in the equation, and the disparity is computed from Rao's quadratic entropy using inter-species distance as shown in the equation.
\begin{equation}
    Heterogeneity = Complexity \times Disparity
\end{equation}
\begin{equation}
    Complexity = E(p) = -\sum_{i=1}^M p_i \times \log p_i
\end{equation}
\begin{equation}
    Disparity = Q(p) =\sum_{i=1}^M \sum_{j=1}^M p_{ij} \times d(i,j)^2
\end{equation}
where $p_i$ is the ratio of a species count to the total population, $d(i,j)$ is inter-species distance between agents $i$ and $j$.

In the current SAR problem, heterogeneity is functionally defined through the difference in knowledge of the agents. This is similar to the computation of heterogeneity from the behavior trees presented in our previous work \cite{parasuraman2020impact}. In the current study, we segregate the robots into four groups, each with the ability to deal with a combination of targets as follows.
\begin{equation}
\centering
\begin{split}
    g_0 = \{\phi\}, & g_1=\{k_R,k_G,k_Y,k_B\},\\ & g_2=\{k_{RG},k_{RY},k_{RB},k_{GY},k_{GB},k_{YB}\}
    \\ & g_3=\{k_{RGY}, k_{RYB}, k_{GYB}\} \\ &g_4=\{k_{RGYB}\} ,
\end{split}
\end{equation}    
where $k_R$ is knowledge of the red target, $k_RG$ is knowledge of red and green, etc.

We define each group type as a species and inter-species distance as the knowledge distance between each set. For e.g., the knowledge distance between $g_0$ and $g_1$ is 1 and between $g_0$ and $g_4$ is 4. This distance estimate is based on the assumption that the knowledge about all target types is similar. If the knowledge for each target type is dissimilar, the groups can be broken further and scored accordingly. Further, for ease of computation, we maintain that the total ability sums to unity. E.g., in the current case, as the knowledge about the targets is similar, we assign 
\begin{equation}
    ks_R=ks_G=ks_Y=ks_B=0.25
\end{equation}

Based on the above-mentioned knowledge scores, we compute the mean knowledge score in the group as follows.

\begin{equation}
    Mean Knowledge Score = \frac{\sum_{i=1}^P ks_i}{P}  
\end{equation}
where $ks_i$ is the total knowledge score of $i^{th}$ agent, $P$ is the total population.

\begin{figure}[t]
    \centering
    \includegraphics[width=0.98\linewidth]{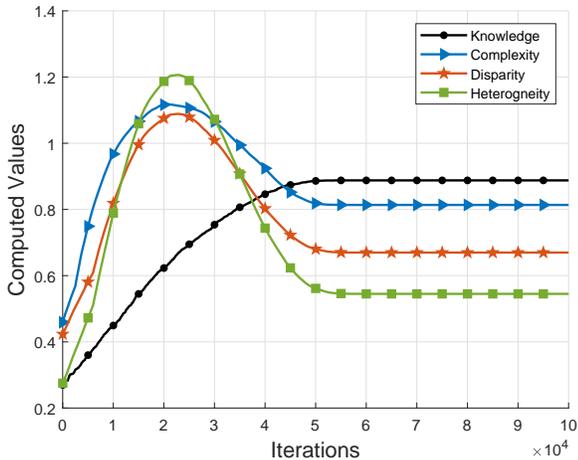}
    \caption{Complexity, Disparity, Heterogeneity and Knowledge score changes over iterations for 10 each of R,G,Y,B knowledge robots in a group (10RGYB) searching for a target combination $(R,G,Y,B)=(100,100,100,100)$}
    \label{fig:Hs10}
    \vspace{-5pt}
\end{figure}

We analyzed the complexity, disparity, and heterogeneity measure based on previously presented equations and compared them against the knowledge factor as shown in Figs.~\ref{fig:Hs391} and \ref{fig:Hs10}. In Fig.~\ref{fig:Hs391}, we present the results obtained for a group with 39 ignorant and one multi-target robot (39I-1M). In Fig.~\ref{fig:Hs10}, we present the results for a group with 40 population size, with members equally distributed with the knowledge to handle the R, G, Y, and B targets (10 each), (10RGYB).

In the 39I-1M group, at the start of the simulations, the system had low complexity as there were 39 homogeneous agents and a high disparity as the knowledge level difference between the ignorant and multi-target robots is high. As robots shared knowledge, more agents moved from lower to higher levels of intelligence. Approximately halfway, while opportunities lasted, both complexity and disparity peaked as the group is now comprised of multiple robots with various levels of intelligence. Finally, the system slowly became homogeneous as all the robots' knowledge levels converged at level 4. The stagnation of heterogeneity beyond 70k iterations is due to the lack of opportunities, which is also evident through the saturation observed in the knowledge factor.

A similar trend can also be observed in the 10RGYB combination, as shown in Fig.~\ref{fig:Hs10}. In contrast to the previous 39I-1M combination, the group starts with slightly higher complexity than disparity as there are four different types of robots but with a comparably lower distinction in knowledge, thus demonstrating lower functional heterogeneity. Trends similar to the 39I-1M composition are observed in complexity, disparity, and knowledge graphs. In both cases, as the knowledge factor saturated, the heterogeneity remained constant, thus, supporting the argument of functional heterogeneity's association with knowledge and opportunities. When more opportunities are provided, when sufficient knowledge is shared, the heterogeneity measure settles at zero as all the agents have the same knowledge factor and thus forming a homogeneous group. This demonstrates the applicability of the KT-BT framework for explicit knowledge sharing tightly integrated with robot control.

\section{Conclusions}
\label{sec:conclusion}
This paper introduced a new framework called KT-BT, which uses behavior trees to transfer knowledge (functional behaviors) between robots through direct communication. This framework can propagate and expand intelligence within a multi-robot and multi-agent system, ultimately achieving homogeneous high-potent knowledge starting from heterogeneous low-potent knowledge spread in individual robots.
We established the rules for a query-response mechanism for knowledge sharing and presented mathematical analysis on knowledge transfer, knowledge spread, and opportunities. We also introduced a \textit{stringBT} grammatical representation of behavior trees to facilitate BT transfer. 

We demonstrated an application of the KT-BT framework on a SAR problem involving a variety of robots that search for various targets and deposit them at their corresponding collection zones. In addition, we developed a unique simulator for conducting studies on knowledge transfer, spread, the effect of knowledge transfer on overall group performance, the effect of opportunity count, and the effect of communication range.
The results demonstrate successful knowledge transfers and improved group performance in various scenarios. 
In our future work, we plan to analyze the KT-BT framework under the contexts of memory-limited computing resources on robots and passive transfer without explicit queries.


\bibliographystyle{IEEEtran}
\bibliography{Mendeley_References.bib,OtherRefs.bib}

\end{document}